\documentclass[11pt,a4paper]{article}
\usepackage[T1]{fontenc}
\usepackage[utf8]{inputenc}
\usepackage{lmodern}
\usepackage[margin=25mm]{geometry}
\usepackage{microtype}
\usepackage{amsmath,amssymb}
\usepackage{graphicx}
\usepackage{booktabs}
\usepackage{cite}
\usepackage{xurl}
\usepackage[hidelinks]{hyperref}
\usepackage[font=small,labelfont=bf]{caption}
\usepackage[section]{placeins}

\newcommand{\R}{\mathbb{R}}
\newcommand{\M}{\mathcal{M}}

\hypersetup{
  pdftitle={GeoTrussRover: Morphological Computation with Contact-Semantic Control Primitives},
  pdfauthor={Muyuan Ma, Yi Zhang, Yang Yang, Xuanyan Zheng, Ruiqi Hu, Boxuan Ke, Zhenyu Chen, Yicong Lin, Xin Hao Yang, Daliang Xiao, Zhinan Hou, Wanhao Niu, Yuan Sun, Yan Yang, Yue Xie},
  pdfsubject={Reconfigurable robotics; morphological computation; variable-geometry truss},
  pdfkeywords={GeoTrussRover, variable-geometry truss, morphological computation, reconfigurable robot, whole-body control}
}

\title{\LARGE\bfseries
GeoTrussRover: Morphological Computation\\
with Contact-Semantic Control Primitives
}

\author{
Muyuan Ma\textsuperscript{*,\textdagger},
Yi Zhang\textsuperscript{\textdagger},
Yang Yang,
Xuanyan Zheng,
Ruiqi Hu,\\
Boxuan Ke,
Zhenyu Chen,
Yicong Lin,
Xin Hao Yang,
Daliang Xiao,\\
Zhinan Hou,
Wanhao Niu,
Yuan Sun,
Yan Yang,
and Yue Xie\\[0.6em]
\small\textsuperscript{\textdagger} Muyuan Ma and Yi Zhang contributed equally to this work.\\
\small\textsuperscript{*} Corresponding author: Muyuan Ma.
}
\date{September 2026}

\begin{document}
\maketitle
\raggedbottom

\begin{abstract}
Reconfigurable robots can change their contact geometry when a fixed body cannot negotiate an obstacle. A variable-geometry truss (VGT) distributes this shape change through a load-bearing structure, but coupling it to a mobile base creates a high-dimensional coordination problem.
GeoTrussRover combines an electrically actuated VGT, a wheeled base, and contact-semantic morphology planning and control.
We solve one source traversal and extract four contact-semantic primitives that describe coordination among 21 members. Physics-constrained projection adapts them to unseen step heights with the same contact topology. When every phase remains feasible, adaptation does not recompute the complete motion. If one phase violates the new physical constraints, only that phase is recomputed. A full-space QP then tracks the adapted motion and corrects member and wheel errors.
For transfer from 0.10\,m to 0.075\,m, the method reduces objective-function evaluations by 63.7\% relative to full recomputation. Contact-phase feasibility analysis covers step heights from 0.10 to 0.46\,m, or 1.08 to 4.97 wheel radii, with the upper value near the theoretical feasible boundary. The electric prototype traverses 2.11 wheel radii.
The resulting low-dimensional representation stores task coordination in a hyper-redundant, load-bearing morphology and reuses it during locomotion.
\end{abstract}

\section{Introduction}

Wheels provide efficient transport on regular terrain, but a vertical step can remove traction, block the chassis, and shift the support region away from the direction of travel.
Wheeled-legged robots address these effects with articulated limbs and whole-body control \cite{Klemm2019Ascento,Bjelonic2020Rolling,Chamorro2024Reinforcement,Lee2024Learning}.
Transformable wheels and wheel-leg mechanisms instead modify the contact geometry itself \cite{Bishop2024Transformable,Lai2025StairClimbing,Lee2025SoftWheel}.
Both approaches show that obstacle traversal depends on contact geometry as well as the motion command.

A VGT applies shape change to the load-bearing body.
Its prismatic members form a closed structural graph that can move the wheels, raise the chassis, alter the support region, and redistribute load without serial multi-axis legs.
This versatility comes with a coordination cost. A few physical task objectives must be converted into compatible commands for many coupled members, and a change in the environment can require the complete constrained path to be solved again.

Step tasks at different heights follow a common contact sequence.
The front wheels engage the obstacle, the structure clears the edge, support transfers forward, the rear wheels recover, and the nominal morphology is restored.
Although the exact member lengths vary with height, the role and ordering of these coordinated changes remain similar.
A solved traversal provides a local coordinate chart for geometrically different tasks that share the same contact topology.

GeoTrussRover exploits this structure in three layers.
First, a constrained source solve converts phase-level task objectives into a sequence of full 21-member configurations.
The consecutive differences form contact-semantic morphology primitives that encode which members should move together.
Second, target-height constraints project those features onto a new contact-conditioned feasible set, so adaptation updates the essential support configurations without recomputing the complete traversal.
Third, a primitive-guided full-space QP (PG-QP) places the projected path in a quadratic tracking objective.
The primitive supplies coordinated nominal motion, while the remaining full-dimensional freedom is used for local pose, support, and contact correction.

This reuse of body coordination follows the idea of morphology-facilitated computation \cite{Pfeifer2006Morphological,Paul2006Morphological,Muller2017What,Sadati2022Embodied}.
The solved shape sequence stores coordination that need not be recovered for every height. A source solve is reused through constrained projection, while the online controller handles residual errors. We measure the benefit through solver evaluations and planning time under the same physical constraints.

\begin{figure}[t]
    \centering
    \includegraphics[width=\linewidth]{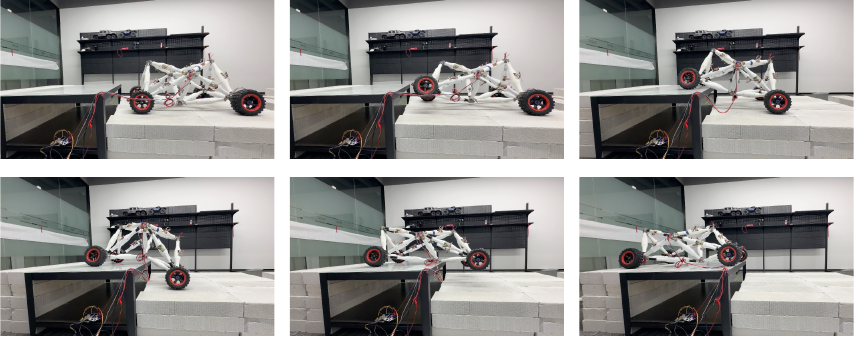}
    \caption{GeoTrussRover traverses a step through body-shape change.}
    \label{fig:physical_sequence}
\end{figure}

Figure~\ref{fig:physical_sequence} shows GeoTrussRover traversing a step through coordinated body-shape change.

This paper makes three contributions:
\begin{itemize}
    \item \textbf{Contact-conditioned manifold representation:} four phase primitives encode a solved 21-member traversal and project it to new step heights under physical constraints.
    \item \textbf{Morphology-facilitated planning and control:} primitive reuse avoids complete replanning or confines recomputation to a mismatched phase, while full-space feedback preserves local corrections.
    \item \textbf{Electromechanical realization:} an electrically actuated, load-bearing wheeled VGT validated through theory, simulation, and physical experiments.
\end{itemize}

\section{Related Work}

\subsection{Morphological Adaptation for Environmental Interaction}

Morphologically adaptive robots alter their body or appendages to match environmental constraints \cite{Sun2023EmbeddedShape}.
Wheel-legged systems combine efficient rolling with articulated terrain interaction, using trajectory optimization, whole-body model predictive control, or learned policies to coordinate wheels, limbs, and body motion \cite{Bjelonic2020Rolling,Bjelonic2021Wholebody,Medeiros2020Trajectory,Lee2024Learning}.
Transformable wheel-leg mechanisms change their contact radius or contact mode for stairs and discontinuous terrain \cite{Bishop2024Transformable,Lai2025StairClimbing,Liu2025RWDDOF}.
The soft deployable airless wheel uses structural deformation to enlarge a compact wheel from 230 to 500\,mm and traverse a 200\,mm obstacle \cite{Lee2025SoftWheel}.
These designs concentrate adaptation in a few limbs or wheels, so the controller has a compact set of morphology variables.

Other robots distribute shape change through the body.
Reconfigurable mobile bases vary width, length, or wheel placement to improve access and stability \cite{Song2022portable,Cheah2022MIRRAX,Karamipour2020Reconfigurable}.
Modular robots compose locomotion modes from repeated units \cite{Yim2000PolyBot,Murata2003MTRAN}, while tensegrity and truss robots exploit coupled structural deformation for rolling and support transfer \cite{Sabelhaus2015SUPERball,Park2019Optimizationbased}.
Embedded shape-morphing modules integrate actuation, sensing, and locking within the body and demonstrate morphology-dependent locomotion modes \cite{Sun2023EmbeddedShape}.
Distributed deformation expands the available motions, but also makes the control coordinates less obvious.

VGT planners address this coupling through support-aware rolling, polygonal configuration-space search, and configuration-space decomposition for collision-free reconfiguration \cite{Park2019Optimizationbased,Park2020Polygonbased,Liu2020FastVTT}.
For shape-shifting variable-stiffness robots, coupled motion-and-morphology planning places pose, morphology, and mode in a common state and coordinates mode-specific MPC through a supervisory state machine \cite{Labazanova2026MotionMorphology}.
These methods provide physically meaningful solutions for a specified environment. Repeating the search for every geometric scale remains costly when the interaction sequence is unchanged.

\subsection{Whole-Body Coordination and Morphology-Facilitated Control}

Whole-body control resolves multiple motion and contact objectives into actuator commands while respecting dynamics and inequality constraints.
In wheeled-legged locomotion, online trajectory optimization and MPC coordinate body motion, wheel rolling, contact forces, and gait sequence \cite{Bjelonic2020Rolling,Bjelonic2021Wholebody,Sun2020Towards}.
Learning-based controllers improve robustness under terrain variation and address difficult-to-model body dynamics \cite{Chamorro2024Reinforcement,Lee2024Learning,Laschi2023LearningSoft}.
Morphology--control co-design is most valuable when a fixed embodiment approaches workspace or obstacle-induced capability limits \cite{Zhang2026CoDesign}.
For a closed truss graph, whole-body coordination must also preserve loop closure and member stroke while maintaining useful contact support.

Movement primitives provide a way to reuse recurring coordination.
Dynamic movement primitives encode attractor dynamics in a low-dimensional parameterization and support temporal and spatial modulation \cite{Ijspeert2013DMP}.
Direct spatial scaling is insufficient for a VGT because new member lengths must still satisfy nonlinear closure, actuator travel, collision clearance, and contact-dependent load constraints.
A truss primitive must instead be adapted on the feasible morphology set.

Morphological computation frames body geometry and material dynamics as part of the information-processing loop \cite{Pfeifer2006Morphological,Paul2006Morphological,Muller2017What,Hauser2023Leveraging}.
Energy has also been used to compare how body, control, and environment shape embodied behavior \cite{Xie2026Energy}.
Metatruss systems group actuators into pneumatic control networks and optimize the grouping and actuation sequence together \cite{Gu2025Metatruss}. This embeds coordination in hardware connectivity. Task-conditioned morphology representations can instead retain independent actuators while storing reusable coordination for constrained target projection.

\section{Method}

\subsection{Electromechanical Variable-Geometry Truss}

GeoTrussRover consists of two octahedral cells joined at a triangular interface.
Spherical joints connect 21 independently actuated telescopic members into a fixed nine-node graph, while four driven wheel modules provide propulsion and differential steering.
The members form a triangulated, load-bearing truss.
Coordinated extension changes body height, wheelbase, lateral span, and attitude without adding serial multi-axis limbs.
Table~\ref{tab:platform} summarizes the platform dimensions and actuator limits.

\begin{figure}[t]
    \centering
    \includegraphics[width=0.96\linewidth]{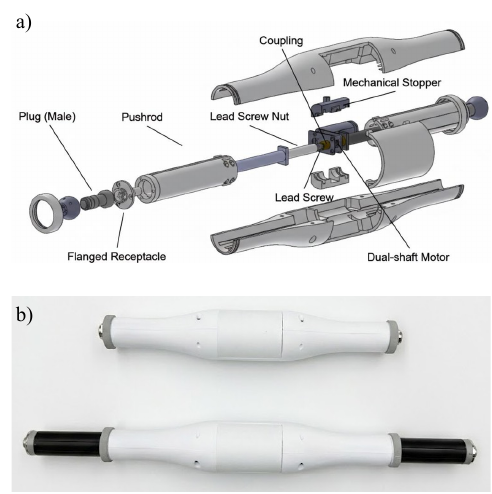}
    \caption{\textbf{Electric VGT member.} (a) Exploded view of the symmetric dual-screw actuator. (b) Retracted and extended prototypes. Opposite-handed screws produce symmetric extension and transfer load through the central housing.}
    \label{fig:actuator_design}
\end{figure}

The custom electric member in Fig.~\ref{fig:actuator_design} converts motor rotation into symmetric telescopic motion.
A dual-shaft motor drives left- and right-handed lead screws through a common coupling, moving both endpoints with respect to the central housing.
Mechanical stops define the stroke, rolling interfaces reduce binding near full extension, and flanged receptacles connect the member to the truss joints.
Encoder-based local control closes the member-position loop, while power and cascaded Controller Area Network communication share a 24\,V bus.
Each module provides a bidirectional load path together with local sensing and low-level control.

\begin{table}[t]
\centering
\caption{GeoTrussRover platform parameters.}
\label{tab:platform}
\setlength{\tabcolsep}{3pt}
\begin{tabular}{@{}lrlr@{}}
\toprule
Quantity & Value & Quantity & Value \\
\midrule
Nodes & 9 & Active members & 21 \\
Driven wheels & 4 & Wheel diameter & 0.185\,m \\
Retracted member & 0.342\,m & Extension ratio & 1.46 \\
Unloaded speed & 7\,mm\,s$^{-1}$ & Rated-load speed & 6.5\,mm\,s$^{-1}$ \\
Bidirectional thrust & 54\,N & Retracted height & 0.496\,m \\
\bottomrule
\end{tabular}
\end{table}

\subsection{Graph Kinematics and Quasi-Static Load Framework}
\begin{figure}[t]
    \centering
    \includegraphics[width=0.7\linewidth]{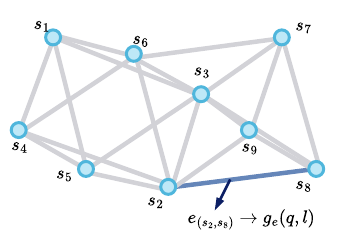}
    \caption{\textbf{Graph of the GeoTrussRover platform.} Stacking all distance constraints $g_e(q,\ell)=0$ gives graph closure.}
    \label{fig:graph_kinematics}
\end{figure}
\begin{figure}[t]
    \centering
    \includegraphics[width=0.99\linewidth]{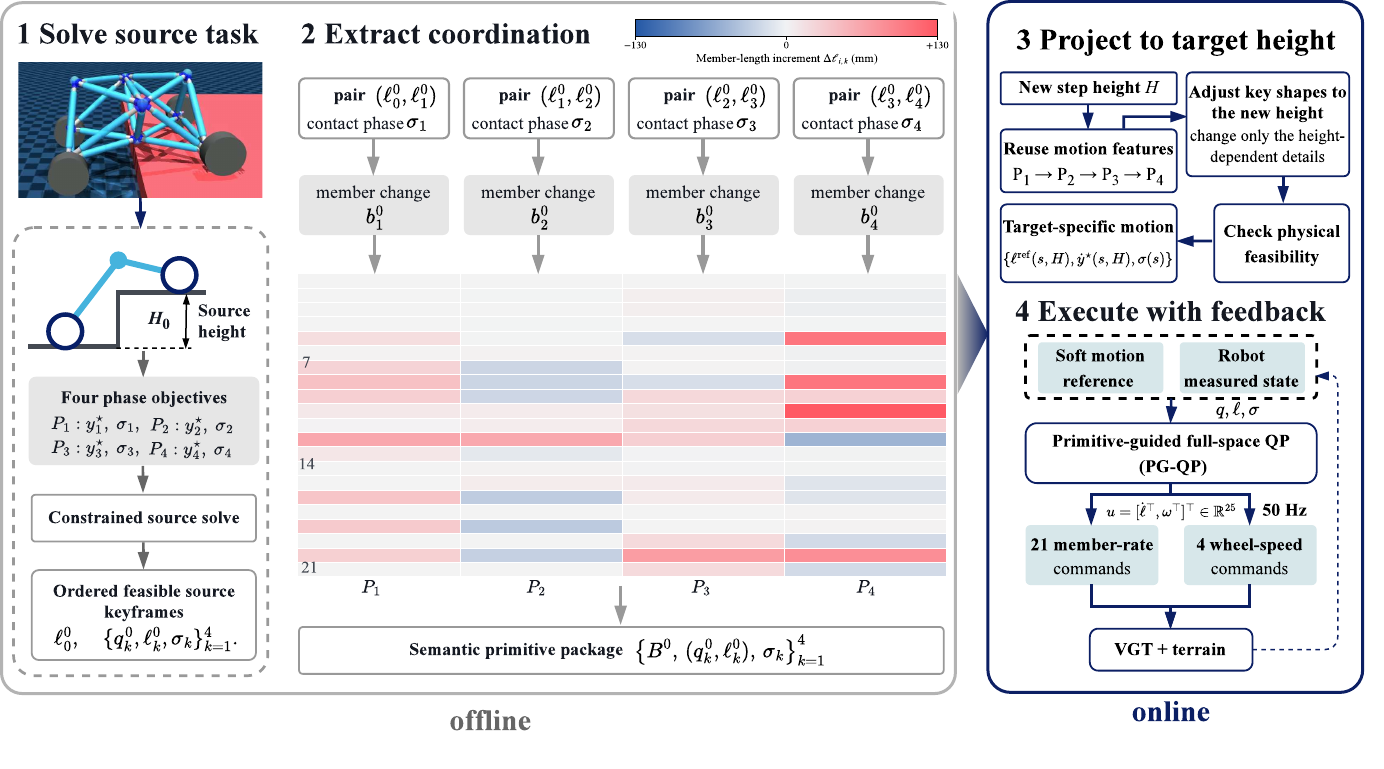}
    \caption{\textbf{Method pipeline.} A source traversal at $H_0$ yields four contact-semantic member-coordination primitives. Physics-constrained projection adapts them to height $H$, and a whole-body QP tracks the reference using 21 member rates and four wheel speeds.}
    \label{fig:system_overview}
\end{figure}

The method has four interfaces, shown in Fig.~\ref{fig:system_overview}.
The fixed graph maps node geometry to compatible member lengths. Contact and equilibrium constraints select a load-bearing morphology branch for each phase. The source primitives carry coordinated shape changes across step heights, and the online controller maps the projected reference and measured state to member and wheel commands.
This separation assigns structural closure to kinematics, support feasibility to the contact model, and execution errors to feedback. One source solve captures the 21-member coordination, target projection adjusts the height-dependent geometry, and full-dimensional feedback retains actuator-level correction.

Let $\mathcal{G}=(\mathcal{V},\mathcal{E})$ denote the fixed truss graph with nodes $\mathcal{V}=\{s_1,\ldots,s_9\}$.
The Cartesian position of $s_i$ is $p_i$, giving the node-position vector $q=[p_1^\top,\ldots,p_9^\top]^\top\in\R^{27}$ and member-length vector $\ell=[\ell_1,\ldots,\ell_{21}]^\top\in\R^{21}$.
Figure~\ref{fig:graph_kinematics} shows the complete topology and highlights member $e_{(s_2,s_8)}$ as one representative graph constraint.

For edge $e=(i,j)$, closure requires
\begin{equation}
g_e(q,\ell)=\tfrac{1}{2}\left(\|p_i-p_j\|_2^2-\ell_e^2\right)=0 .
\label{eq:closure}
\end{equation}
Stacking Eq.~\eqref{eq:closure} gives $g(q,\ell)=0$ and the velocity relation
\begin{equation}
\dot\ell=R(q)\dot q,
\label{eq:rigidity}
\end{equation}
where $R(q)\in\R^{21\times27}$ is the normalized rigidity matrix.
At a regular configuration, $\operatorname{rank}R=21$ and the six-dimensional null space contains rigid-body motions.

For an active support phase $\sigma$, member axial forces $f$ and wheel--terrain forces $\lambda$ satisfy the quasi-static balance
\begin{equation}
\begin{aligned}
R(q)^\top f+J_{c,\sigma}(q)^\top\lambda+\beta w_g(q)&=0,\\
\lambda\in\mathcal{K}_\sigma,\qquad |f_e|&\leq F_e^{\max}.
\end{aligned}
\label{eq:static_balance}
\end{equation}
where $J_{c,\sigma}$ is the active-contact Jacobian, $\mathcal{K}_\sigma$ is the linearized friction cone, and $w_g$ is the nominal nodal gravity wrench.
The largest feasible gravity scale defines the load multiplier $\beta^\star$ used below.
Equations~\eqref{eq:closure}--\eqref{eq:static_balance} connect planning and control. Planning returns closed, load-admissible references, local servos track the member lengths, and the 50\,Hz QP uses tangent kinematics and measured support to correct robot--terrain deviations.

\subsection{Contact-Conditioned Morphology Manifold}

The task is to traverse a vertical step of height $H$ by reshaping the load-bearing graph while maintaining a feasible sequence of wheel supports.
We decompose the traversal into front engagement, edge clearance, support transfer, rear recovery, and morphology restoration because each event changes the active contact constraints and structural load path.
These constraints place the admissible shapes on contact-dependent branches of the closed-graph configuration space rather than in an unconstrained Euclidean member-length space.
We represent each branch by a contact-conditioned morphology manifold.
For phase $\sigma$, equalities $c_{\sigma,H}(q)=0$ set wheel and task coordinates, while inequalities $d_{\sigma,H}(q)\geq0$ enforce envelope clearance, support, and symmetry tolerances.
The usable morphology set is
\begin{equation}
\begin{split}
\M_{\sigma,H}=\{(q,\ell)\mid {}&g(q,\ell)=0,\ c_{\sigma,H}(q)=0,\\
&d_{\sigma,H}(q)\geq0,\ \ell^-\leq\ell\leq\ell^+,\\
&\operatorname{rank}R(q)=21,\ \beta^\star(q)\geq1\}.
\end{split}
\label{eq:morphology_set}
\end{equation}
The quasi-static multiplier $\beta^\star$ scales the nominal gravitational wrench under linearized friction cones and member-force limits.
Where the equality Jacobian has constant rank, its local solution forms a manifold and the inequalities select the physically usable region.
Each contact phase defines a local branch rather than a single global surface.

\subsection{Contact-Semantic Primitive Coordinates}

At a source height $H_0$, the designer specifies four phase outcomes in the robot body frame: front-right seating, front-left seating, edge-clearance elevation, and wheelbase contraction for rear recovery.
A constrained solve chooses the complete graph configuration for each outcome:
\begin{equation}
\begin{aligned}
(q_k^0,\ell_k^0)=\arg\min_{q,\ell}\quad&
\|W_y[y_k(q)-y_k^\star(H_0)]\|_2^2\\
&+\lambda\|W_\ell(\ell-\ell_{\mathrm{nom}})\|_2^2\\
\mathrm{s.t.}\quad&(q,\ell)\in\M_{\sigma_k,H_0}.
\end{aligned}
\label{eq:source_solve}
\end{equation}
The phase targets state what the body must achieve; Eq.~\eqref{eq:source_solve} determines how the 21 members coordinate to achieve it.

Set $\ell_0^0=\ell_{\mathrm{nom}}$ and define the consecutive differences
\begin{equation}
b_k^0=\ell_k^0-\ell_{k-1}^0,
\qquad
B^0=[b_1^0,\ldots,b_4^0]\in\R^{21\times4}.
\label{eq:primitive_extract}
\end{equation}
Each column of $B^0$ describes the coordinated 21-member change for one contact phase.
Cumulative columns reconstruct the source endpoints, while continuation knots $\gamma_k^0(s)$ provide a smooth reference within each phase.
The four primitives parameterize the source path, not the complete feasible morphology set.

\subsection{Physics-Projected Transfer Across Height}

At target height $H$, the source endpoint initializes the optimization and the phase objective changes from $y_k^\star(H_0)$ to $y_k^\star(H)$.
The adapted endpoint solves
\begin{equation}
\begin{aligned}
\min_{q,\ell}\quad&\|W_q(q-q_k^0)\|_2^2\\
&+\lambda\|W_\ell(\ell-\ell_k^0)\|_2^2\\
\mathrm{s.t.}\quad&(q,\ell)\in\M_{\sigma_k,H},\\
&\|y_k(q)-y_k^\star(H)\|_\infty\leq\varepsilon_y .
\end{aligned}
\label{eq:projection}
\end{equation}
Projection can adjust every member to recover closure, stroke, clearance, and support at the new height. Its initialization already contains the source phase coordination.
The target primitive is again the difference between consecutive projected endpoints.
A full solve starts from the nominal morphology, whereas projection starts near the stored source path and corrects its height-dependent mismatch.

\begin{figure}[t]
    \centering
    \includegraphics[width=0.81\linewidth]{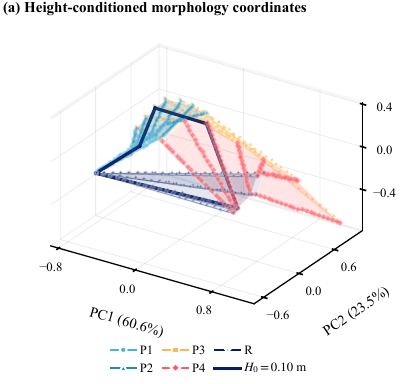}
    \vspace{-1.5mm}
    \includegraphics[width=\linewidth]{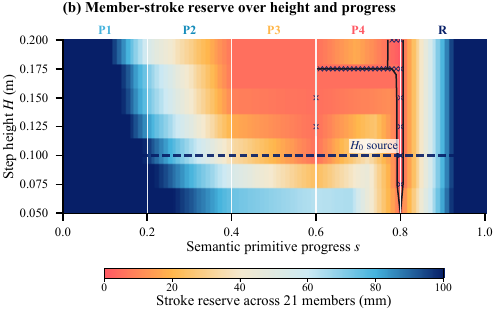}
    \caption{\textbf{Height-conditioned morphology family and stroke reserve.} (a) Feasible 21-member configurations embedded in the first three principal coordinates. Colors identify P1--P4 and restoration, and each polyline follows one step height. (b) Minimum remaining stroke across all members versus step height and semantic progress.}
    \label{fig:morphology_family}
\end{figure}

Figure~\ref{fig:morphology_family} visualizes the resulting height-conditioned morphology family and the member-stroke reserve available along the semantic sequence.

\begin{figure}[t]
    \centering
    \includegraphics[width=0.8\linewidth]{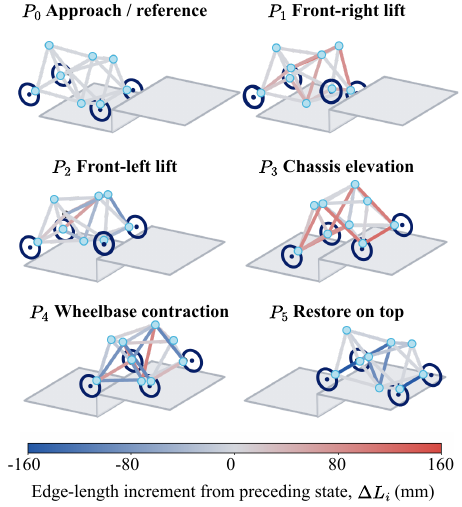}
    \caption{\textbf{Contact-semantic morphology sequence at $H=0.20$\,m.} Blue/red members contract/extend from the preceding configuration.}
    \label{fig:primitives}
\end{figure}

Figure~\ref{fig:primitives} shows the semantic sequence at one target height.
Color gives the signed member-length change between consecutive configurations, directly exposing the coordination reused by projection.

\subsection{Primitive-Guided Full-Space QP}

The online PG-QP stage in Fig.~\ref{fig:system_overview} receives the projected primitive reference and the measured robot state.
Its decision variable retains all member rates and wheel speeds,
\begin{equation}
u=[\dot\ell^\top,\omega^\top]^\top\in\R^{25}.
\label{eq:control}
\end{equation}
Let $y=h(q)$ collect center-of-mass motion, orientation, axle position and height, and lateral span.
Truss kinematics, the active support model, and wheel rolling give
\begin{equation}
\dot y=G_\sigma(q)u,
\qquad G_\sigma(q)=J_y(q)K_\sigma(q).
\label{eq:global_map}
\end{equation}
At 50\,Hz, the controller solves
\begin{equation}
\begin{aligned}
\min_{u,\xi}\;&
\|G_\sigma u-\dot y^\star\|_Q^2
+\|\ell+\Delta t\dot\ell-\ell^{\mathrm{ref}}(s,H)\|_P^2\\
&+\|u-u^-\|_S^2+\rho\|\xi\|_2^2\\
\mathrm{s.t.}\;&\ell^-\leq\ell+\Delta t\dot\ell\leq\ell^+,\\
&|\dot\ell_i|\leq6.5\ \mathrm{mm\,s}^{-1},\quad
|\omega_j|\leq4.034\ \mathrm{rad\,s}^{-1},\\
&A_\sigma(q)u\leq b_\sigma(q)+\xi,\quad \xi\geq0 .
\end{aligned}
\label{eq:qp}
\end{equation}
The primitive-tracking term supplies the coordinated task direction.
Keeping $u\in\R^{25}$, rather than imposing $\dot\ell=B\dot\phi$, leaves the whole-body QP free to correct measured pose, load, and contact errors.
The low-dimensional coordinates act as a soft guide without removing actuator-level feedback directions.

\section{Experiments and Results}

We evaluate the roles of shape actuation and wheel contact, primitive-guided coordination, transfer to new heights, and physical implementation.

\subsection{Why Morphology and Contact Matter}

We analyze five support phases over $H\in\{0.10,0.15,0.20,0.25,0.46\}$\,m using wheel drive alone, shape actuation alone, and contact-augmented actuation.
The contact-augmented map incorporates the active wheel--terrain kinematics into the truss motion map, so the 21 member rates and four wheel speeds generate task motion through the same support geometry.
We evaluate rigidity, stroke reserve, task-achievement ratio $\alpha^\star$, and quasi-static load multiplier $\beta^\star$ at every configuration. Table~\ref{tab:theory} reports the mean task-achievement ratio across the tested heights.

\begin{table}[t]
\centering
\caption{Mean task-achievement ratio across tested heights.}
\label{tab:theory}
\begin{tabular}{@{}lccc@{}}
\toprule
Contact phase & Wheel only & Shape only & Augmented \\
\midrule
All wheels on ground & 1.0 & 0.32 & 1.0 \\
Front wheels on wall & 0.53 & 0.049 & 0.82 \\
Front wheels on top & 1.0 & 0.32 & 1.0 \\
Rear wheels on wall & 0.41 & 0.049 & 0.74 \\
All wheels on top & 1.0 & 0.32 & 1.0 \\
\bottomrule
\end{tabular}
\end{table}

At wall support, neither wheel drive nor shape actuation is sufficient by itself. Wheel drive cannot reshape the support geometry, and shape actuation produces little motion along the wall without wheel rolling.
The augmented map combines both inputs. Relative to wheel drive alone, task authority rises by about 1.5 times at front-wall contact and 1.8 times at rear-wall contact. It is more than 15 times the shape-only value at both phases.
Across the sampled configurations, the minimum $\beta^\star$ is 1.72 and the projected 0.46\,m configuration retains 9.36\,mm of member travel. These values make the sampled support phases load-admissible within the tested model.

Along the executed 0.10 and 0.075\,m paths, $\alpha^\star$ stays above 0.32 and $\beta^\star$ above 1.31. Both minima occur during rear-left lift, where the successful trajectories also have their lowest rear-axle load. Rear recovery is the main support-transfer bottleneck.
The denser sweep in Fig.~\ref{fig:morphology_family}, covering $H\in[0.05,0.20]$\,m, shows little change during front seating but larger corrections and lower stroke reserve during elevation and rear recovery. Direct replay or geometric scaling cannot satisfy these constraints throughout the sequence. This motivates extracting and adapting phase-wise primitives from the 21-member motion.

\subsection{Primitive Guidance in Full-Space Whole-Body Execution}

The four phase primitives compress recurring coordination among 21 coupled members. We test a primitive-subspace controller (PS-QP), a reference-free full-space controller (RF-QP), and primitive-guided full-space control (PG-QP) on the 0.10\,m source task, the projected 0.075\,m task, and a perturbed 0.075\,m task. The ratio $E_\perp/E$ is the fraction of member-command energy outside the primitive subspace, which measures correction beyond the nominal coordination.

\begin{table}[t]
\centering
\caption{Primitive use and constrained whole-body execution.}
\label{tab:primitive_execution}
\footnotesize
\begin{tabular}{@{}lcccc@{}}
\toprule
Controller & 0.10\,m & 0.075\,m & Perturbed & $E_\perp/E$ \\
\midrule
PS-QP, hard subspace & Fail & Fail & Fail & 0.00 \\
RF-QP, no primitive & Fail & Fail & Fail & 0.57 \\
PG-QP, soft guidance & Pass & Pass & Pass & 0.72 \\
\bottomrule
\end{tabular}
\end{table}

\begin{figure}[t]
    \centering
    \includegraphics[width=0.9\linewidth]{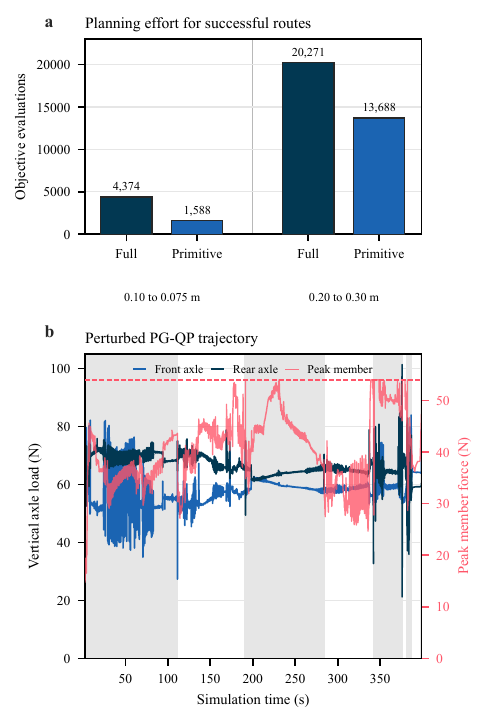}
    \caption{\textbf{Planning transfer and closed-loop execution.} (a) Objective evaluations for Full and Primitive planning in high-to-low and low-to-high transfer. (b) Axle loads and peak member force during the perturbed PG-QP traversal. The dashed line marks the member-force limit.}
    \label{fig:minimal_main_results}
\end{figure}

Only PG-QP completes all three conditions in Table~\ref{tab:primitive_execution}. PS-QP removes correction directions needed at contact transitions. RF-QP retains those directions but reaches sustained rate saturation during the first wheel lift because it lacks phase coordination. In the perturbed run, PG-QP places 72\% of its member-command energy outside the primitive subspace. The primitive supplies the task direction without blocking local correction. PG-QP solves in about 4.17\,ms, compared with 16.21\,ms for RF-QP. The force constraint in Fig.~\ref{fig:minimal_main_results}b holds the peak member force at the stated limit during contact transitions.

\subsection{Single-Source Transfer to Unseen Step Heights}

We test primitive reuse from a higher step to a lower step and in the opposite direction. Full recomputes every target phase. Primitive projects the stored source coordination and recomputes a phase only when it falls outside the transferable region.

For transfer from 0.10\,m to 0.075\,m, the complete primitive maps directly to the new feasible manifold. Full requires 4374 objective evaluations, whereas Primitive requires 1588. This reduces the evaluation count by 63.7\% and synthesis time by 66.5\%. Both produce complete references, and Primitive completes the PG-QP traversal. Projection preserves the source coordination without repeating the global search.

Transfer from 0.20\,m to 0.30\,m places greater demands on the morphology. Applying the 0.20\,m solution unchanged and projecting the complete path both stall during rear recovery. Figure~\ref{fig:morphology_family}a shows the same pattern: P4 has the largest displacement as height changes. Recomputing load transfer and rear recovery while reusing the earlier primitives produces a successful 0.30\,m route. Primitive uses 13,688 evaluations, 32.5\% fewer than the 20,271 used by Full. Under the same enhanced 250\,N condition, both routes complete the traversal and maintain four-wheel support on top for 3.33\,s. Primitive reuse reduces planning for directly transferable tasks and confines new computation to the phase that no longer transfers.

\subsection{Simulation and Physical Demonstrations}

\begin{figure}[t]
    \centering
    \includegraphics[width=\linewidth]{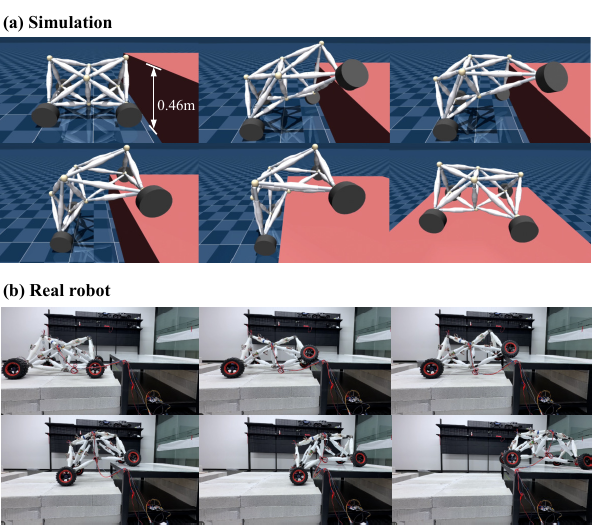}
    \caption{\textbf{Simulation and physical step traversal.} (a) Isaac Sim sequence on a 0.46\,m step. (b) Electrically actuated prototype traversing a 0.195\,m step.}
    \label{fig:simulation_physical}
\end{figure}

We evaluated GeoTrussRover in MuJoCo, Isaac Sim, and hardware. MuJoCo provides the closed-loop results reported above. In Isaac Sim, the robot completes the semantic sequence on a 0.46\,m step near the theoretical morphology limit. The electric prototype traverses a 0.195\,m step, more than twice the wheel radius, through dynamic shape change (Fig.~\ref{fig:simulation_physical}). These results test the morphology range in simulation and confirm that the mechanism can execute the sequence on hardware.

\section{Discussion}

\subsection{Morphology-Facilitated Motion Control}
The contact-augmented analysis treats morphology as an actuated part of the controller. Wheel drive cannot reshape the support geometry at wall contact, whereas the 21 VGT members change wheel placement and load paths. Combining shape actuation with wheel contact raises the task-achievement ratio from 0.53 to 0.82 at front contact and from 0.41 to 0.74 at rear contact. The primitive stores the resulting coordination for projection and PG-QP, reducing endpoint computation by 63.7\%. The VGT contributes both geometric reach and part of the coordination used during online control.

\subsection{Dynamic Base and Shape Control Coupling in the Sim-to-Real Gap}
Whole-body shape control is expressed in the non-inertial frame of the accelerating base. Changes in wheelbase, axle orientation, mass distribution, and wheel load remain coupled through contact. Increasing friction uniformly can lock a wheel against the vertical face. In the 0.15\,m diagnostic, rear contact transition occurs only when ground release and step traction are separated and all four wheels cooperate. More torque, lower mass, a larger lift reference, or rear-wheel drive alone does not produce the same transition. The run uses uncalibrated contact separation and exceeds the member-speed limit, so it supports this coupling explanation rather than a validated controller result. The gravity ramp and settling period also reduce closed-chain impact and establish preload. Further work requires actuator identification, calibrated CAD collision geometry, and load-aware whole-body control.

\section{Conclusion}

GeoTrussRover turns one solved traversal into reusable contact-semantic primitives. They compress 21-member motion into phase coordination, physics projection adapts that coordination to a target-height manifold, and PG-QP retains actuator-level correction. Primitive transfer reduces endpoint evaluations by 63.7\% from 0.10\,m to 0.075\,m. From 0.20\,m to 0.30\,m, local recomputation of rear recovery reduces evaluations by 32.5\%. The feasibility analysis covers 1.08 to 4.97 wheel radii and shows that shape actuation and wheel contact are complementary. The electric prototype traverses 2.11 wheel radii through whole-body shape change. The load-bearing, hyper-redundant body stores reusable task structure without giving up local feedback during locomotion.

\section*{ACKNOWLEDGMENT}
OpenAI GPT-5.6 Sol assisted with software-architecture planning, automation of repeatability tests for the completed simulation models, language editing, manuscript typesetting, and conversion of Type 3 fonts to submission-compliant vector graphics.

\bibliographystyle{IEEEtran}
\bibliography{refs}

\end{document}